\def\year{2022}\relax
\documentclass[letterpaper]{article} 
\usepackage{aaai22}  
\usepackage{times}  
\usepackage{helvet}  
\usepackage{courier}  
\usepackage[hyphens]{url}  
\usepackage{multirow}
\usepackage{graphicx} 
\usepackage{natbib}  
\usepackage{caption} 
\DeclareCaptionStyle{ruled}{labelfont=normalfont,labelsep=colon,strut=off} 
\usepackage{algorithm}
\usepackage{amsmath,amsfonts}
\usepackage{algorithmic}
\usepackage{subfig}
\usepackage{gensymb}
\usepackage{siunitx}

\usepackage{amsmath,amsfonts,bm}

\def\eqref#1{equation~\ref{#1}}

\def\1{\bm{1}}

\DeclareMathAlphabet{\mathsfit}{\encodingdefault}{\sfdefault}{m}{sl}
\SetMathAlphabet{\mathsfit}{bold}{\encodingdefault}{\sfdefault}{bx}{n}

\DeclareMathOperator*{\argmax}{arg\,max}

\usepackage{booktabs} 
\usepackage{newfloat}
\usepackage{listings}
\floatstyle{ruled}
\newfloat{listing}{tb}{lst}{}
\floatname{listing}{Listing}
\title{Revisiting Latent-Space  Interpolation \\ via a Quantitative Evaluation Framework}
\author{
Lu Mi$^1$\thanks{~~Equal contribution. \\ ~~Corresponding to \{lumi, tianxing\}@mit.edu}, Tianxing He$^1$\footnotemark[1], Core Francisco Park$^2$, Hao Wang$^3$, Yue Wang$^1$, Nir Shavit$^1$
}
\affiliations{$^1$MIT, $^2$Harvard University, $^3$Rutgers University
}

\usepackage{bibentry}

\begin{document}

\maketitle

\begin{abstract}
Latent-space interpolation is commonly used to demonstrate the generalization ability of deep latent variable models. Various algorithms have been proposed to calculate the best trajectory between two encodings in the latent space. In this work, we show how data labeled with semantically continuous attributes can be utilized to conduct a quantitative evaluation of latent-space interpolation algorithms, for variational autoencoders. Our framework can be used to complement the standard qualitative comparison, and also enables evaluation for domains (such as graph) in which the visualization is difficult. Interestingly, our experiments reveal that the superiority of interpolation algorithms could be domain-dependent. While normalised interpolation works best for the image domain, spherical linear interpolation achieves the best performance in the graph domain. Next, we propose a simple-yet-effective method to restrict the latent space via a bottleneck structure in the encoder. We find that all interpolation algorithms evaluated in this work can benefit from this restriction. Finally, we conduct interpolation-aware training with the labeled attributes, and show that this explicit supervision can improve the interpolation performance.
\end{abstract}

\section{Introduction}
\label{sec:intro}

Generative latent variable models, especially variational autoencoders (VAEs) \citep{Kingma2014}, and generative adversarial networks (GANs) \citep{Goodfellow14gan}, have been gaining huge research interest in recent years. For both VAEs and GANs, a decoder (or generator, in GAN terminology) is learned to generate data samples $x$ from a latent variable $z$, which is sampled from a prior distribution. For VAE, the decoder is usually probabilistic (i.e., $p(x|z)$), while for GAN the generator is a one-to-one mapping function.

Latent-space interpolation has been widely used for thw evaluation of generative latent variable models, as a way of demonstrating that a generative model generalizes well, instead of simply memorizing the training examples. It is usually implemented in an unsupervised and qualitative manner: A trajectory is formed between two latent encodings sampled from the prior distribution. Then, the decoded samples (e.g., images) from the latent encodings along that trajectory are qualitatively examined. Intuitively, a good interpolation trajectory should be decoded into meaningful outputs exhibiting a gradual transition. Meanwhile, it should also reflect the internal structure of the data.

Linear interpolation, where the selected trajectory is simply the shortest path in Euclidean space, has been most widely used in the literature \citep{Kingma2014,radford2015unsupervised}. Despite its simplicity, multiple recent studies \citep{white16spherical,agustsson2018optimal,lesniak2018distributioninterpolation} point out that linear interpolation introduces a \textit{mismatch} between the distribution of the interpolated points and the original prior distribution. And such a mismatch could potentially prevent the interpolation from faithfully reflecting the inner structure of the generative model. From this viewpoint, several interpolation algorithms have been proposed to eliminate this distribution mismatch (to be reviewed in the background section). 

However, the unsupervised nature of deep generative model training makes a principled comparison between different interpolation algorithms challenging. Most of the recent studies apply these algorithms to generate a trajectory between two points sampled from the latent space, and qualitatively compare the outputs (usually in the form of images) decoded from successive latent encodings on the trajectory. This is also due to the ``discrete'' nature of the commonly used datasets such as LSUN \citep{fisher15lsun} or CelebA \citep{liu15celeba}: Given two data samples, there are no ground-truth reference samples that are supposed to be semantically ``between'' them.  Moreover, in some domains such as a graph, it is difficult to visualize the decoded samples, making a qualitative comparison almost impossible.

In this work, we attempt to compare different interpolation algorithms in a  quantitative evaluation framework, \textbf{by utilizing datasets with semantically continuous attributes}. Here is one example: Consider that we train a VAE with an image dataset containing views of a 3-D object from different angles. Naturally, we would expect the latent encodings to capture the variations of the angles. If we interpolate the latent encodings of 30-degree and 90-degree views of the same object, it is natural to expect that the decoded image from the mid-point interpolation to be close to the 60-degree view. Further, we can synthesize new views of the object from any angle in between, by changing the interpolation weight. In this way, we can evaluate the interpolation algorithm by comparing the decoded image to the ground-truth image with standard metrics.

Our experiments cover image-domain and graph-domain VAE models. We briefly summarize the key messages from our experiments as follows: Our evaluation reveals that the superiority of interpolation algorithms could be domain-dependent. Normalised interpolation works best in the image domain, while spherical linear interpolation achieves the best performance in the graph domain. Next, we propose a simple-yet-effective method to restrict the latent space via a bottleneck structure in the encoder. And we report that the interpolation algorithms evaluated in this work can all benefit from this restriction. Finally, we conduct interpolation-aware training with the labeled attributes, and we show that this explicit supervision can boost the interpolation performance. 

\vspace{-0.1cm}
\section{Background}
\label{sec:background}
\vspace{-0.05cm}

In this section, we review the framework of variational autoencoders, and introduce three existing interpolation algorithms. As a generative latent variable model, VAE \citep{Kingma2014}, adopts a two-step generation process: (1) First, the $D$-dimensional latent variable $z$ is sampled from a fixed prior distribution $p(z)$ (e.g., standard Gaussian). (2) Then, a decoder network $p_\theta(x|z)$ maps $z$ into the data space. $\theta$ denotes the trainable parameters in the decoder. 

The existence of the latent variable makes direct maximum likelihood estimation (MLE) training difficult. Instead, the evidence lower bound (ELBO) loss is adopted, where an inference network $q_\phi(z|x)$ is introduced:
\begin{equation}
\small
\label{eq:std_elbo}
    \mathcal{L}_\text{ELBO}=-\mathbb{E}_{z\sim q_\phi(z|x)}\log p_\theta(x|z) + D_\text{KL}(q_\phi(z|x)||p(z)),
\end{equation}
where $D_\text{KL}$ refers to Kullback-Leibler divergence (definition given in Appendix A). Note that we will also refer to the inference work as the encoder in the following text.

It can be derived that $-\mathcal{L}_\text{ELBO}=\log p_\theta(x) -D_\text{KL}(q_\phi(z|x)||p_\theta(z|x))$. Therefore, $-\mathcal{L}_\text{ELBO}$ is a lower bound of the log-likelihood of $x$. And during the minimization of $\mathcal{L}_\text{ELBO}$, the inference network $q_\phi(z|x)$ is implicitly trained to approximate the true posterior $p_\theta(z|x)$.

Latent-space interpolation has been used to qualitatively examine how well the model generalizes. It forms a trajectory between two latent encodings $z_1$ and $z_2$, which are independently sampled from the prior distribution. Following the notations from \citet{lesniak2018distributioninterpolation}, we formulate interpolation algorithms as a function $f$:
\begin{equation}
\small
    f: \mathbb{R}^D \times \mathbb{R}^D \times [0,1] \ni (z_1, z_2, \lambda) \mapsto z \in \mathbb{R}^D,
\end{equation}
where $\lambda$ is referred to as the interpolation weight. Next, we review several existing interpolation algorithms.

\textit{Linear interpolation} is the most widely used interpolation algorithm, which simply forms a straight line between $z_1$ and $z_2$. We formulate it  below:
\begin{equation}
\small
    f^\text{linear}(z_1, z_2, \lambda) = (1-\lambda) z_1 + \lambda z_2.
\end{equation}

\textit{Spherical linear interpolation}~\citep{white16spherical}  treats the interpolation as a great circle path on a $D$-dimensional hypersphere. It utilizes a formula from \citet{shoemake85animatingrotate}:
\begin{equation}
\small
    f^\text{slerp}(z_1, z_2, \lambda) = \frac{\text{sin}[(1-\lambda)\Omega]}{\text{sin}(\Omega)} z_1 + \frac{\text{sin}[\lambda \Omega]}{\text{sin}(\Omega)} z_2,
\end{equation}
where $\Omega$ is the angle between $z_1$ and $z_2$.

\textit{Normalised interpolation} \citep{agustsson2018optimal} is based on optimal transport maps. It adapts the latent-space interpolation operations, so that the resulting interpolated points would match the prior distribution. Based on linear interpolation with Gaussian prior, the formulation can be derived as: 
\begin{equation}
\small
    f^\text{norm}(z_1, z_2, \lambda) = \frac{(1-\lambda)z_1 + \lambda z_2}{\sqrt{(1-\lambda)^2+\lambda^2}}.
\end{equation}

\vspace{-0.1cm}
\section{Evaluation Methodology}
\label{sec:framework}
\vspace{-0.1cm}

In this section, we introduce our evaluation framework for interpolation algorithms. As discussed in the introduction section, commonly used datasets for VAE training do not have reference samples for the evaluation of interpolation. Take the popular CelebA dataset \citep{liu15celeba} as an example, given two images of human faces, it is difficult to locate reference face images that are supposed to be ``between'' these two faces. 

Motivated by this problem, in this work we utilize datasets with \textit{temporal} or \textit{spatial} attributes. For example, we will use citation networks that evolve with time, and 2-D rendering of 3-D objects from different angles. Formally, we denote our dataset as $\{\langle t,x_t \rangle\}$, where each data sample $x$ comes with a \textbf{semantically continuous attribute} $t \in \mathbb{R}$. For VAE baseline training, we will ignore the $t$ labels, and just train the model in an unsupervised manner with $\{x\}$. Therefore, we expect the latent encodings to capture the variance related to attribute $t$.

Our proposed evaluation procedure is motivated by an \textit{application} point of view. Given $x_{t_1}$ and $x_{t_3}$, the generative latent variable model and the  interpolation algorithm enable us to \textbf{synthesize} $x_{t_2}$ at any $t_2$ between $t_1$ and $t_3$. We now describe it in more details below.

To evaluate a given interpolation algorithm $f$, we randomly select a number of triples in the test set $\{(x_{t_1},x_{t_2},x_{t_3})\}$ with $t_1 < t_2 < t_3$. We then infer $(\hat{z}_{t_1}, \hat{z}_{t_2}, \hat{z}_{t_3})$ by taking $\hat{z}_t:=\argmax_z q_\phi(z|x_t)$.\footnote{As discussed in the background section, $q_\phi(z|x)$ is implicitly trained to approximate the true posterior $p_\theta(z|x)$.} For the VAEs used in this work, this is equivalent to simply taking the outputed mean vector from the inference model. 

An estimation of $z_{t_2}$ can be obtained by applying the interpolation algorithm $z^\text{inter}_{t_2}:=f(\hat{z}_{t_1}, \hat{z}_{t_3}, \frac{t_3-t_2}{t_3-t_1})$. Then we feed $z^\text{inter}_{t_2}$ to the decoder to generate $x^\text{inter}_{t_2}:=\argmax_x p_\theta(x|z^\text{inter}_{t_2})$. Finally, by measuring the distance between $x^\text{inter}_{t_2}$ and $x_{t_2}$, or the distance between $z^\text{inter}_{t_2}$ and $\hat{z}_{t_2}$ with standard metrics, we can quantify the performance of the interpolation algorithm $f$.

\vspace{-0.1cm}
\section{Model Formulations}
\vspace{-0.1cm}

In this section, we first formulate the standard VAE models considered in this work. We then discuss how we adjust these models or the objective functions for better interpolation performance.

\vspace{-0.05cm}
\subsection{Variational Autoencoders}
\label{sec:model_vae}
\vspace{-0.05cm}

To prepare baseline models, we use a standard convolutional neural network (CNN)-based VAE model for the image-domain data, and a graph convolutional network (GCN)-based VAE model for the graph-domain data. 

\vspace{-0.1cm}
\paragraph{CNN-based VAE} For the image-domain VAE, we use a popular implementation of ``vanilla'' VAE from \citet{Subramanian2020}. The encoder contains 6 layers of CNN \citep{Krizhevsky12cnnimage}.\footnote{The hidden dimensions of the CNN is [16, 32, 64, 128, 256, 512].} Each CNN layer is followed by a batch normalization layer and a LeakyReLU activation. Then two fully-connected output layers output the $D$-dimensional mean and standard deviation vector for the posterior Gaussian distribution $q_\phi(z|x)$. The decoder is composed of 6 deconvolution layers, followed by a convolution layer with a Tanh activation for the final output. We use standard Gaussian as the prior distribution. The standard ELBO loss (Equation \ref{eq:std_elbo}) is adopted.

\vspace{-0.1cm}
\paragraph{Graph Variational Autoencoders (GVAEs)}  We denote a directed graph as $\mathcal{G}=(\mathcal{V}, \mathcal{E})$, with $N=|\mathcal{V}|$ nodes and edge set $\mathcal{E}$. From $\mathcal{E}$, we construct its adjacency matrix $A$. Each node $i$ is assigned a $D$-dimensional latent encoding $z_i$, summarized in an $N \times D$ matrix $Z$.\footnote{When we apply spherical linear interpolation, we will concatenate all $z_i$ and treat it as a long vector.} For node features, we simply use one-hot representation, giving a $N \times N$ feature matrix $X$. 

We develop a version of GVAE \citep{kipf17gcn} for directed graphs. Our GVAE model consists of a two-layer graph convolutional network (GCN)  encoder, and a decoder consisted of an MLP and followed by a linear product. For the inference model, we first encode the graph via the GCN and get $ \{ ( e^\mu_i , e^\sigma_i )\}=\text{GCN}(X, A) $, where $( e^\mu_i , e^\sigma_i )$ is a pair of $D$-dimensional embeddings for each node $i$. We refer readers to \citet{Kipf2016VariationalGA} for details about the GCN. We now formulate inference model\footnote{Our code is based on \url{https://github.com/DaehanKim/vgae_pytorch}.} as follows:

\begin{equation}
\small
    q_\phi(Z|X,A) = \prod^N_{i=1} \mathcal{N}(z_i|\text{MLP}_{\text{enc}_\mu}(e^\mu_i), \exp (e^\sigma_i)),
\end{equation}
where $\mathcal{N}$ denotes the Gaussian distribution. To grant the model  more flexibility, a two-layer multi-layer perceptron (MLP) with a hidden dimension of $D$ is added for the mean-output branch.

For decoding, we first pass $z_i$ through a 2-layer MLP and get $z'_i:=\text{ MLP}_\text{dec}(z_i)$. To model the directness of the graph, we do a simple half-half separation of $z'_i$ into two $\frac{D}{2}$-dimensional vectors $z'_{i1}$ and $z'_{i2}$. Finally, the generative model is formulated as a simple inner product with a trainable bias term $b$ to control the graph sparsity:
\begin{equation}
\small
    p_\theta(A|Z)=\prod^N_{i=1}\prod^N_{j=1}p_\theta(A_{ij}|z_i,z_j),
\end{equation}
where $p_\theta(A_{ij}|z_i,z_j)=\sigma({z'}^\top_{i1} z'_{j2} + b)$, and $\sigma(\cdot)$ is the logistic sigmoid function.

The GVAE is then optimized with the ELBO loss:
\begin{equation}
\scriptsize
    \mathcal{L}^\text{GVAE}_\text{ELBO} =  - \mathbb{E}_{q_\phi(Z|X,A)}[\log p_\theta(A|Z)] + D_\text{KL}(q_\phi(Z|X,A)||p(Z)),
\end{equation}
where we use standard Gaussian for the prior $p(Z)=\prod^N_{i=1} \mathcal{N}(Z_i|0, I)$.

\vspace{-0.1cm}
\subsection{Restricting the Latent Space}
\label{sec:model_lowrank}
\vspace{-0.1cm}
Intuitively, it could be easier for interpolation algorithms to locate the right $z_{t_2}$ when the latent space is simpler. 

In our experiments, we explore two intuitive ways to restrict the latent space. (1) We directly shrink the latent dimension $D$ (the hidden dimension in the encoder and decoder model is not changed). (2) We enforce a bottleneck structure for the posterior-mean branch of the encoder. To be more specific, suppose the original weight parameter of the final linear layer is a matrix $W$ of size $H \times D$, where $H$ is the final hidden dimension in the encoder. Given a rank $R$, we replace $W$ by $W^R_1W^R_2$, where $W^R_1$ and $W^R_2$ are $H \times R$ and $R \times D$ matrices, respectively. In this way, the mean of the Gaussian posterior $q_\phi(z|x)$ will be restricted to the (at most) $R$-dimensional linear space spanned by the row vectors of $W^R_2$. The intuition is that by restricting the power of the inference network $q_\phi(z|x)$, we expect that the true posterior distribution $p_\theta(z|x)$ would be encouraged to be simpler as well. 

\vspace{-0.1cm}
\subsection{Interpolation-Aware Training}
\label{sec:model_explicittrain}
\vspace{-0.1cm}

With the additional attributes, it is attractive to conduct \textit{interpolation-aware} training (IAT) in a supervised manner, where we encourage the model to be consistent with the interpolation algorithm we choose. Take a step further, we can directly parametrize the interpolation function via a function approximator such as an MLP. Below, we take the CNN-based VAE as an example, and formulate different variants of IAT objective functions.

We first consider adding a penalty to the model to encourage it to be consistent with a chosen interpolation algorithm $f$. This can be done either on latent-encoding level, or decoder-output level:
\begin{equation}
\small
\begin{split}
    &\mathcal{L}_\text{IAT}^\text{latent} = \mathbb{E}_{(x_{t_1}, x_{t_2}, x_{t_3})} ||\hat{z}_{t_2} - f(\hat{z}_{t_1}, \hat{z}_{t_3}, \frac{t_3-t_2}{t_3-t_1})||^2_2,  \\
    &\mathcal{L}_\text{IAT}^\text{decode} = \mathbb{E}_{(x_{t_1}, x_{t_2}, x_{t_3})} \log p_\theta(x_{t_2}|z=f(\hat{z}_{t_1}, \hat{z}_{t_3}, \frac{t_3-t_2}{t_3-t_1})),
\end{split}
\end{equation}
where $\hat{z}_t$ is the output mean from the encoder model $q_\phi(z|x_t)$.\footnote{We do not do stop-gradient for $\hat{z}_t$, therefore the encoder will also be updated by $\mathcal{L}_\text{IAT}$.} 

Further, we attempt to parametrize the interpolation function via a 3-layer MLP with a hidden dimension of $D$ and a ReLU activation:
\begin{equation}
\small
    \begin{split}
    &\mathcal{L}_\text{IAT}^\text{MLP+latent} = \mathbb{E}_{(x_{t_1}, x_{t_2}, x_{t_3})} ||\hat{z}_{t_2} - \text{MLP}^\text{inter}(\hat{z}_{t_1}, \hat{z}_{t_3}, z^\text{inter}_{t_2})||^2_2, \\
     &\mathcal{L}_\text{IAT}^\text{MLP+dec} = \mathbb{E}_{(x_{t_1}, x_{t_2}, x_{t_3})} \log p_\theta(x_{t_2}|z=\text{MLP}^\text{inter}(\hat{z}_{t_1}, \hat{z}_{t_3}, z^\text{inter}_{t_2})), 
    \end{split}
\end{equation}
where $z^\text{inter}_{t_2}=f(\hat{z}_{t_1}, \hat{z}_{t_3}, \frac{t_3-t_2}{t_3-t_1})$. The input to the MLP is a concatenation of $\hat{z}_{t_1}, \hat{z}_{t_3}, \hat{z}^\text{inter}_{t_2}$, where the information about the interpolation weight is contained in $\hat{z}^\text{inter}_{t_2}$. Note that the  $\mathcal{L}^\text{MLP+latent}_\text{IAT}$ variant can be regarded as an informal upper bound of what's the best interpolation algorithms can do.

The final joint objective is a weighted combination of $\mathcal{L}_\text{ELBO}$ and $\mathcal{L}_\text{IAT}$:
\begin{equation}
    \mathcal{L}_\text{joint-IAT}=\mathcal{L}_\text{ELBO} + \lambda_\text{IAT} \mathcal{L}_\text{IAT},
\end{equation}
where $\lambda_\text{IAT}$ controls the weight of $\mathcal{L}_\text{IAT}$.

\vspace{-0.1cm}
\section{Datasets and Tasks}
\label{sec:data}
\vspace{-0.1cm}

Three datasets are adopted for our experiments: ShapeNet and CO3D for the image domain, and Citation Network for the graph domain. We describe them below:

\vspace{-0.1cm}
\paragraph{ShapeNet}  The dataset we introduce to evaluate interpolation quantitively in the image domain is generated from ShapeNet~\citep{shapenet2015}. We construct a dataset including $20$ chairs of different styles, and each object is rendered with 60 different horizontally rotation degrees that are randomly selected between $0\si{\degree}$ and $180\si{\degree}$. We use a fixed light position and a fixed distance between the camera and objects during rendering. The renderer applied here is composed of a rasterizer and a shader, which is implemented by Pytorch3D~\citep{ravi2020pytorch3d}. For each chair, we use $30$ randomly selected angles as the training set, $15$ angles as the validation set, and $15$ angles as the test set. Note that the training and validation sets are only used for IAT. We design an image interpolation task as the following: We randomly select a chair and a triplet of angles $(x_{t_1},x_{t_2},x_{t_3})$, and the model is then asked to synthesize $x_{t_2}$ given information of $x_{t_1},x_{t_3}$ and $t_2$, via latent-space interpolation. Note that for evaluation or IAT, we only interpolate between the same chair of different rotation angles. From the test set, $2000$ sampled triplets are used for evaluation.

\paragraph{CO3D} We also adopt a real-world image dataset: Common Objects in 3D (CO3D)~\citep{reizenstein2021common}. We create a dataset including $434$ chairs of different styles. Each object (chair) has about $100$ consecutive views. These views constitute a natural rotation or translation of the object. We regard the index of the view as $t$. For each object, we use $60$ randomly selected views as the training set, $20$ views as the validation set, and $20$ views as the test set. From the test set, $21700$ sampled triplets are used for evaluation.

\vspace{-0.1cm}
\paragraph{Citation Network} The dataset we introduce to perform the evaluation on the graph domain is generated from the high energy physics citation network~\citep{leskovec2005graphs}. To limit the total number of nodes in consideration, we first choose the most-cited paper as the center core, and then only consider papers that cite the center core. This process creates a directed graph with $661$ nodes, where each node represents a paper. Then, using the submission time of these papers, we create a temporal graph with $200$ time stamps spanning $10$ years. The adjacency matrix $A_t$ grows as papers cite each other. We randomly select $100$ time stamps as the training set, $50$ stamps as the validation set, and $50$ stamps as the test set. We design a graph interpolation task as the following: We first randomly select a triplet of time stamps $({t_1},{t_2},{t_3})$. Given information of $A_{t_1},A_{t_3}$ and $t_2$, the model is asked to predict $A_{t_2}$ via latent-space interpolation. $2000$ sampled triplets are used for evaluation.

\vspace{-0.1cm}
\section{Experiments}
\label{sec:exp}
\vspace{-0.1cm}

Our experiments can be roughly divided into three sections: (1) We use our evaluation framework to compare existing interpolation algorithms, for the unsupervised baseline model. (2) We explore applying a low-dimensional latent space, or a low-rank encoder to restrict the latent space, and test whether the interpolation performance can benefit from this restriction. (3) Finally, we utilize the labeled attributes and conduct interpolation-aware training.

\vspace{-0.1cm}
\subsection{Evaluation of Interpolation Algorithms}
\label{sec:exp_evalcompare}
\vspace{-0.1cm}

To train the VAE models, we use stochastic gradient descent with the Adam optimizer \citep{kingma15adam}. We stop the training iterations when the ELBO loss has converged. Hyper-parameters such as learning rate are tuned based on the validation set performance. For the baseline models of ShapeNet and CitationNet, we use latent dimension $D=512$. For CO3D we use $D=4096$ as it gives better performance.  More details about our implementation are deferred to Appendix B.

The metrics we use to evaluate on ShapeNet in image domain is mean square error (MSE) and \textit{structure similarity index measure} (SSIM)~\citep{hore2010image} between the interpolated image ${x}^\text{inter}_{t_2}$ and the ground truth $x_{t_2}$. For CO3D, we use MSE and \textit{peak signal-to-noise ratio} (PSNR) ~\citep{hore2010image}. For Citation Network, we report with binary cross-entropy (BCE) and the \textit{edge intersection over union} (eIoU) \citep{jaccard12} metric for edge prediction between the adjacency matrices of interpolated graph and reference ground-truth graph. On the other hand, we also quantify the distance between the interpolated encoding ${z}^\text{inter}_{t_2}$ and the inferred encoding $\hat{z}_{t_2}$ using MSE and cosine similarity (denoted by cos-sim).

\begin{table*}[t]
\vspace{-0.1cm}
\small
\centering
\addtolength{\tabcolsep}{-1.9pt}

\begin{tabular}{r|ccc|r|ccc|r|ccc}
\toprule
\multicolumn{4}{c|}{\textbf{ShapeNet}}              & \multicolumn{4}{c|}{\textbf{CO3D}}              & \multicolumn{4}{c}{\textbf{Citation Network}}                                 \\ \midrule
\textbf{Metrics}                                                & $f^\text{linear}$ & $f^\text{slerp}$ & $f^\text{norm}$ & \textbf{Metrics}                                                & $f^\text{linear}$ & $f^\text{slerp}$ & $f^\text{norm}$ & \textbf{Metrics}                                                & $f^\text{linear}$ & $f^\text{slerp}$ & $f^\text{norm}$ \\ \midrule
$\text{MSE}({x}^\text{inter}_{t_2},x_{t_2}) \downarrow$               & 0.133             & 0.151            & \textbf{0.131}  & $\text{MSE}({x}^\text{inter}_{t_2},x_{t_2}) \downarrow$               & 0.274             & 0.226            & \textbf{0.174} & $\text{BCE}({x}^\text{inter}_{t_2},x_{t_2})\downarrow$ &0.950 &\textbf{0.932} &0.938\\
$\text{SSIM}({x}^\text{inter}_{t_2},x_{t_2})\uparrow$                                                   & 0.733             & 0.721            &  \textbf{0.736}  & $\text{PSNR}({x}^\text{inter}_{t_2},x_{t_2})\uparrow$                                                   & 13.31            & 13.65       & \textbf{14.33}    & $\text{eIoU}({x}^\text{inter}_{t_2},x_{t_2})\uparrow$                                                 &0.562 &\textbf{0.590} &0.578          \\ \midrule
$\text{MSE}({z}^\text{inter}_{t_2},\hat{z}_{t_2})\downarrow$         & 0.078             & 0.021            & \textbf{0.015}  &  $\text{MSE}({z}^\text{inter}_{t_2},\hat{z}_{t_2})\downarrow$         & 0.044             & 0.006          & \textbf{0.004}  & $\text{MSE}({z}^\text{inter}_{t_2},\hat{z}_{t_2})\downarrow$         & \textbf{0.009}             & 0.012  & 0.012  \\
$\text{cos-sim}({z}^\text{inter}_{t_2},\hat{z}_{t_2})\downarrow$ & 0.493             & \textbf{0.381}   & 0.502  & $\text{cos-sim}({z}^\text{inter}_{t_2},\hat{z}_{t_2})\downarrow$ & 0.434             &\textbf{0.290}  & 0.436           & $\text{cos-sim}({z}^\text{inter}_{t_2},\hat{z}_{t_2})\downarrow$ & 0.736             & \textbf{0.715}   & 0.736           \\ \bottomrule
\end{tabular}
\vspace{0.2cm}
\caption{A quantitative comparison of different interpolation algorithms. Our evaluation reveals that the superiority of interpolation algorithms could be domain-dependent. While normalised interpolation works best in the image domain, spherical linear interpolation achieves the best performance in the graph domain.
}
\vspace{-0.3cm}
\label{tab:interalg_compare}
\end{table*}

\begin{figure*}[t]
\begin{center}
    \includegraphics[width=1\textwidth]{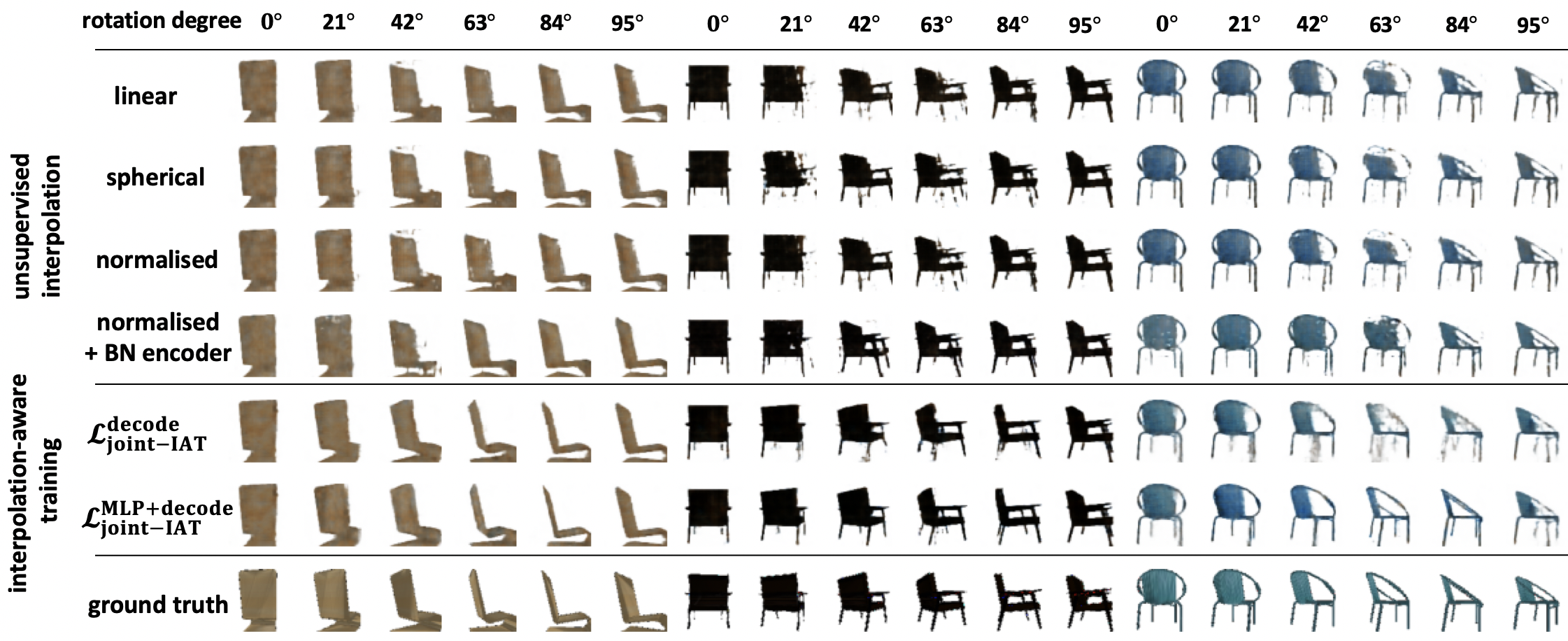}
\end{center}
    \caption{ShapeNet: Qualitative evaluations of different interpolation algorithms including unsupervised baseline and interpolation-aware training.  The ``BN encoder'' refers to the rank-8 bottleneck encoder. The quality gap between different interpolation algorithms for the unsupervised baseline is not very clear.}
    \label{fig:main_qualititave}
    \vspace{-0.1cm}
\end{figure*}

We compare different interpolation algorithms for the baseline VAE models with metrics described above, and the results are shown in Table \ref{tab:interalg_compare}. For the performance w.r.t. $x^\text{inter}_{t_2}$, normalised interpolation performs best both on ShapeNet and CO3D. However, for Citation Network, spherical linear interpolation outperforms the other two algorithms. While this agrees with previous works in that a more sophisticated interpolation algorithm could indeed bring performance gain, it also suggests that the superiority of interpolation algorithms could be \textbf{domain dependent} (note that previous works have been focusing on the image domain). 

For the performance w.r.t. $z^\text{inter}_{t_2}$, interestingly, we observe that for all datasets, spherical linear interpolation achieves the best cosine-similarity score, while normalised interpolation achieves the best MSE score on ShapeNet and CO3D. Motivated by this, we attempt to combine the ideas from both sides as \textit{normalised spherical interpolation} (formulated in Appendix A). However, this combination does not add performance gain, and we omit the results here.

In the upper part of Figure \ref{fig:main_qualititave}, we show a qualitative comparison of the ShapeNet data. While in some cases we do observe that normalised interpolation gives better decoded images than linear interpolation, the quality gap is not clear enough. Our quantitative metrics, on the other hand, give complementary information when the qualitative comparison falls short.

\vspace{-0.1cm}
\subsection{Interpolation in Restricted Latent Space}
\label{sec:exp_lowdim}
\begin{figure*}[t]
\vspace{-10pt}
  \centering
  \subfloat[ShapeNet]{\includegraphics[width=0.33\linewidth]{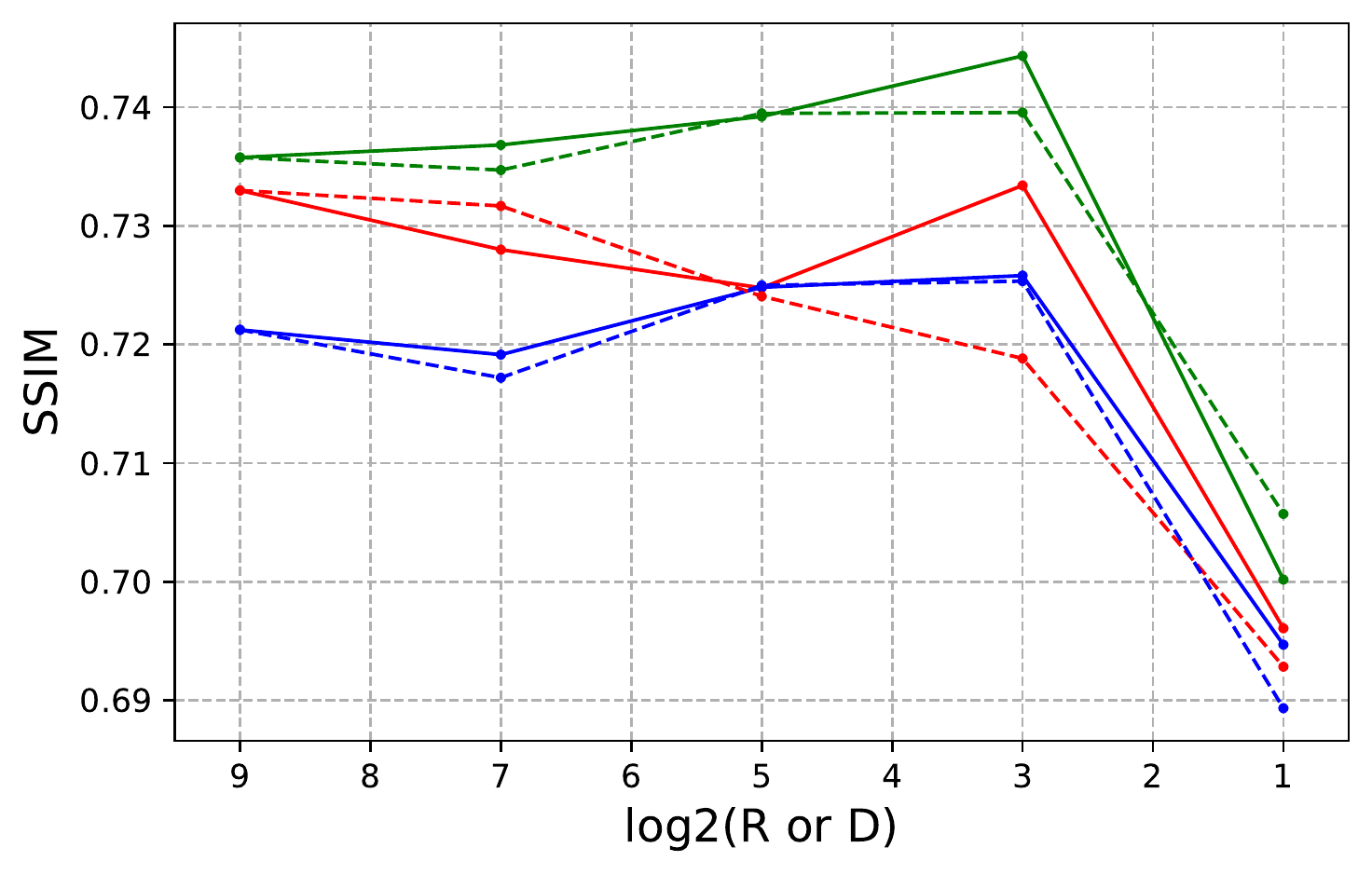}}
  \subfloat[Citation Network]{\includegraphics[width=0.33\linewidth]{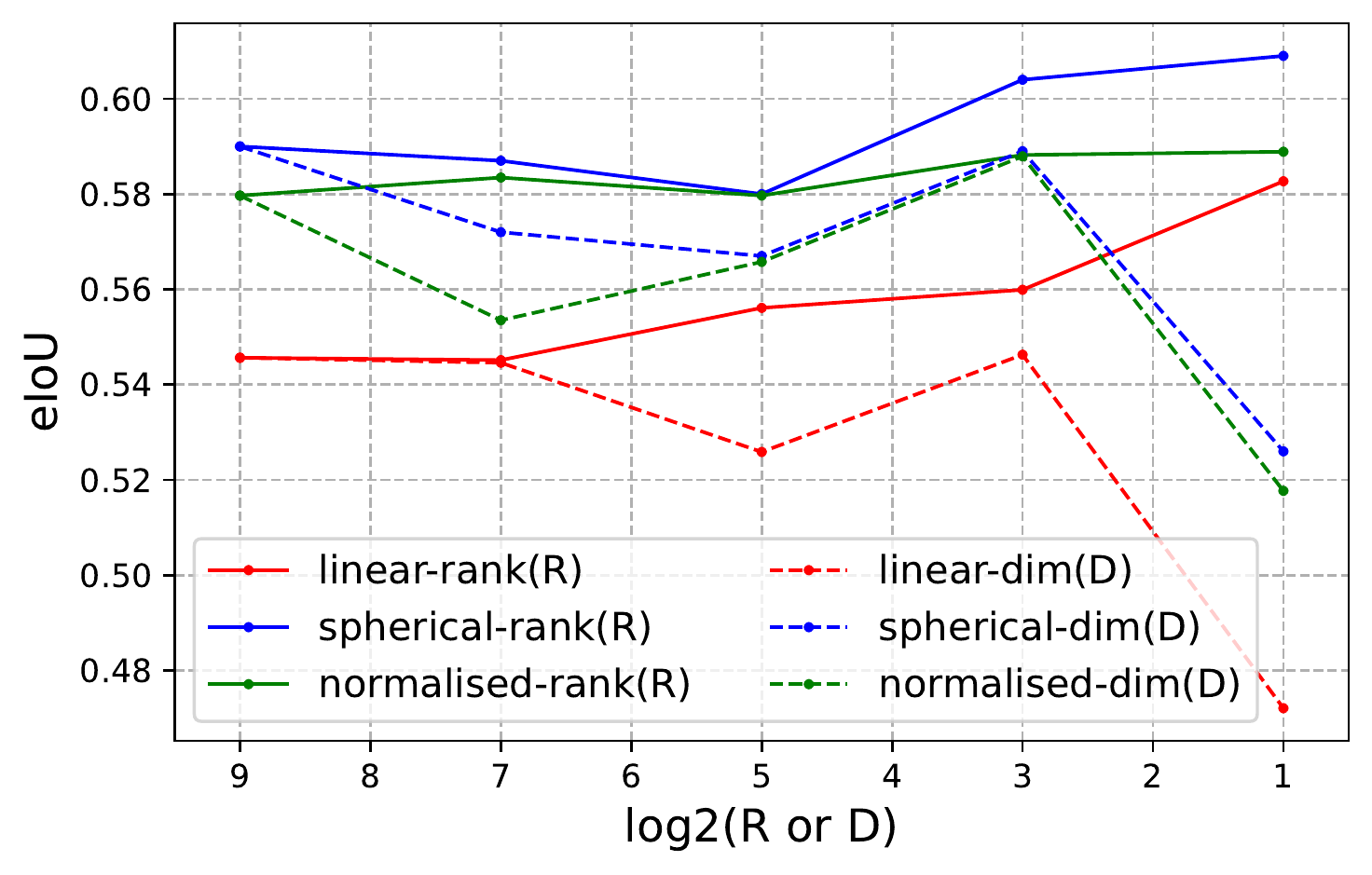}}
  \subfloat[CO3D]{\includegraphics[width=0.33\linewidth]{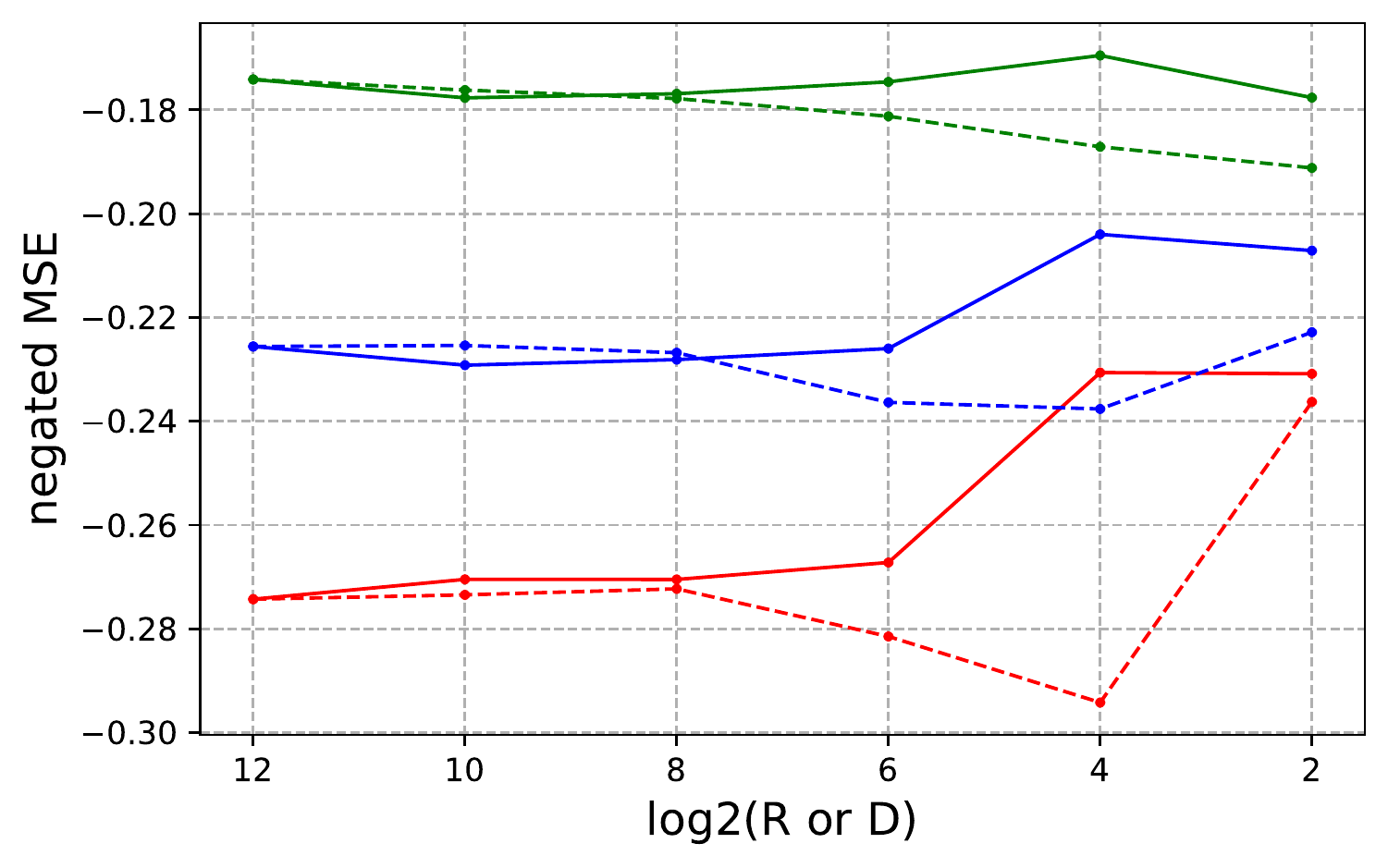}}
  \caption{Interpolation performance with low-dimensional ($D$) latent space, or with low-rank ($R$) encoder. The common legend is shown in the middle figure. In most cases, the low-rank encoder has better performance.}
  \vspace{-0.2cm}
  \label{fig:main_rankdim}
\end{figure*}

\vspace{-0.1cm}
As described in the formulation section, we test two intuitive ways to restrict the latent space: (1) We directly shrink the latent dimension $D$. (2) We adopt a low-rank ($R$) structure to the final linear transform in the encoder. We show the performance of different interpolation algorithms in Figure \ref{fig:main_rankdim}, where we tune $D$ or $R$ in log-scale, and all other hyper-parameters are kept the same with the baseline model.

It is shown that the low-rank encoder gives better performance than shrinking the latent dimension $D$ for all three interpolation algorithms. Especially in the Citation Network dataset, the rank-2 encoder achieves the best performance for different interpolation algorithms. This matches our intuition that interpolation would benefit from a simpler latent space. Consistent with Table \ref{tab:interalg_compare}, normalised interpolation performs best for ShapetNet and CO3D, and spherical linear interpolation performs best for Citation Network. 

We report the detailed measurements in Table \ref{tab:rankdim_metric}. Surprisingly, we observe a misalignment between the performance w.r.t. $z^\text{inter}_{t_2}$ and the performance w.r.t. $x^\text{inter}_{t_2}$. For example, for the ShapeNet dataset, the best-performing (w.r.t. SSIM) rank-8 model has relatively poor MSE or cosine-similarity scores for $z^\text{inter}_{t_2}$. This could be due to the fact that the linear space $\hat{z}_t$ stays in is changing with $R$ or $D$, and therefore these numbers are not directly comparable. 

\begin{table*}[t]
\small
\centering
\footnotesize
\addtolength{\tabcolsep}{-2.0pt}

\begin{tabular}{r|ccccc|ccccc}
\toprule
\textbf{Dataset}                                       & \multicolumn{5}{c|}{\textbf{Latent Dimension}}                               & \multicolumn{5}{c}{\textbf{Encoder Rank}}                                        \\ \midrule
\textbf{ShapeNet ($f^\text{norm}$)}                    & \textbf{D512} & \textbf{D128} & \textbf{D32} & \textbf{D8}     & \textbf{D2} & \textbf{R512} & \textbf{R128}  & \textbf{R32}   & \textbf{R8}    & \textbf{R2}    \\ \midrule
$\text{MSE}({x}^\text{inter}_{t_2},x_{t_2})\downarrow$           & 0.131         & 0.136         & 0.130        & 0.131           & 0.148       & 0.131         & 0.134          & 0.132          & 0.130          & \textbf{0.127} \\$\text{SSIM}({x}^\text{inter}_{t_2},x_{t_2})\uparrow$                                                  & 0.736         & 0.735         & 0.739        & 0.740           & 0.706       & 0.736         & 0.737          & 0.739          & \textbf{0.744} & 0.700          \\ \midrule
$\text{MSE}({z}^\text{inter}_{t_2},\hat{z}_{t_2})\downarrow$     & 0.015         & 0.061         & 0.304        & 0.842           & 0.931       & 0.015         & 0.019          & \textbf{0.014} & 0.018          & 0.015          \\
$\text{cos-sim}({z}^\text{inter}_{t_2},\hat{z}_{t_2})\downarrow$ & 0.502         & 0.433         & 0.402        & 0.474           & 0.604       & 0.502         & \textbf{0.426} & 0.490          & 0.480          & 0.693          \\ \midrule
\textbf{Citation Network ($f^\text{slerp}$)}            & \textbf{D512} & \textbf{D128} & \textbf{D32} & \textbf{D8}     & \textbf{D2} & \textbf{R512} & \textbf{R128}  & \textbf{R32}   & \textbf{R8}    & \textbf{R2}    \\ \midrule
$\text{BCE}({x}^\text{inter}_{t_2},x_{t_2})\downarrow$   &0.932 &0.944 &0.972 &0.988 &0.893 &0.932 &0.924 &0.977
&0.945 &\textbf{0.866}\\
 
$\text{eIoU}({x}^\text{inter}_{t_2},x_{t_2})\uparrow$   &0.590 &0.572
&0.567 &0.589 &0.526 &0.590
&0.587 &0.580 &0.604 &\textbf{0.609}
 \\ \midrule
$\text{MSE}({z}^\text{inter}_{t_2},\hat{z}_{t_2})\downarrow$ & 0.012 &1.586 &1.044 &0.050 &11.784 &0.012 &0.009 &0.007 &\textbf{0.006} &1.270\\
$\text{cos-sim}({z}^\text{inter}_{t_2},\hat{z}_{t_2})\downarrow$ &0.715 &0.723 &0.732 &0.734 &0.731 &0.715 &\textbf{0.714}
&0.726 &0.719 &0.724 \\ \bottomrule
\end{tabular}
\vspace{0.2cm}
\caption{Interpolation performance with various metrics for low-dimensional ($D$) latent space, or with low-rank ($R$) encoder. For ShapeNet, we observe a steady SSIM improvement from R512 to R8. Due to lack of space, we defer results with CO3D to Appendix C.}

\label{tab:rankdim_metric}
\end{table*}

\subsection{Interpolation-Aware Training}
\label{sec:exp_intertrain}
\vspace{-0.1cm}

So far, the labeled attributes have been used to evaluate the interpolation algorithms, while the training remains unsupervised. As discussed in the formulation section,  we apply several variants of interpolation-aware training to the baseline model, and the results are shown in Table \ref{tab:main_iat}. Note that for ShapeNet/CO3D we use the normalised interpolation, while for Citation Network we use spherical linear interpolation.

As expected, IAT greatly boosts the interpolation performance comparing to the unsupervised baseline. Especially, the variants where the interpolation MLP are introduced achieve stronger performance for most metrics, showing the potential of a parametrized interpolation function. Note that the performance of the  $\mathcal{L}^\text{MLP+latent}_\text{IAT}$ variant can be regarded as an informal upper bound of interpolation algorithms for unsupervised training, and the gap suggests that there could still be room for the interpolation algorithms to improve. 

We also observe that the decoder variants give better performance than the latent-encoding variant. This is as expected because in these cases the decoder is directly optimized to predict $x_{t_2}$. In Appendix C, we provide a study of how the performance improves as the number of labeled data grows. 
We show qualitative examples for the decoder variants of IAT in Figure \ref{fig:main_qualititave}. Comparing to the unsupervised models, the decoded chairs from IAT are rotated to the correct angle. Consistent with the scores in Table \ref{tab:main_iat}, the $\mathcal{L}_\text{MLP+decode}$ variant generates the most sharp and clear images.

\begin{table*}[t]
\addtolength{\tabcolsep}{-3.5pt}
\centering
\footnotesize
\begin{tabular}{c|ccccc|c|ccccc}
\toprule

\multicolumn{6}{c|}{\textbf{ShapeNet ($f^\text{norm}$)}} & \multicolumn{6}{c}{\textbf{Citation Network ($f^\text{slerp}$)}} \\ \midrule
\textbf{IAT} & \multicolumn{1}{c}{\textbf{N/A}} & \multicolumn{1}{c}{\textbf{lat.}} & \multicolumn{1}{c}{\textbf{dec.}} & \multicolumn{1}{c}{\textbf{\begin{tabular}[c]{@{}c@{}}mlp \\ + lat.\end{tabular}}} & \multicolumn{1}{c|}{\textbf{\begin{tabular}[c]{@{}c@{}}mlp \\ + dec.\end{tabular}}} & \multicolumn{1}{c|}{\textbf{IAT}} & \textbf{N/A} & \textbf{lat.} & \textbf{dec.} & \textbf{\begin{tabular}[c]{@{}c@{}}mlp \\ + lat.\end{tabular}} & \textbf{\begin{tabular}[c]{@{}c@{}}mlp \\ + dec.\end{tabular}} \\ \midrule
$\text{MSE}({x}^\text{inter}_{t_2},x_{t_2})\downarrow$ & 0.131 & 0.127 & 0.056 & 0.059 & \textbf{0.047} &\multicolumn{1}{c|}{$\text{BCE}({x}^\text{inter}_{t_2},x_{t_2})\downarrow$} &0.932  &0.071  &\textbf{0.004}  &0.056  &0.039 \\
$\text{SSIM}({x}^\text{inter}_{t_2},x_{t_2})\uparrow$ & 0.736 & 0.742 & 0.815 & 0.837 & \textbf{0.859} & \multicolumn{1}{c|}{$\text{eIoU}({x}^\text{inter}_{t_2},x_{t_2})\uparrow$} & 0.590 & 0.605 & 0.736 & 0.682 & \textbf{0.741} \\ 
\bottomrule
\end{tabular}
\caption{Results of interpolation-aware training (IAT). ``lat.'' refers to $\mathcal{L}^\text{latent}_\text{IAT}$, and ``dec.'' refers to $\mathcal{L}^\text{decode}_\text{IAT}$. Results on CO3D are deferred to Appendix C. The supervision provided by the labeled attributes greatly improve the interpolation performance, comparing to the unsupervised baseline (marked by ``N/A'').}
\vspace{-0.2cm}
\label{tab:main_iat}
\end{table*}

\vspace{-0.1cm}
\section{Discussion and Limitations}
\label{sec:discuss}
\vspace{-0.1cm}

We devote this section to discuss the limitations of this work. Our evaluation is motivated by an application point of view, where we apply the interpolation algorithms to output a latent code that can decode to a data sample of a certain attribute value. However, due to the unsupervised nature of standard VAE training, and the fact that existing interpolation algorithms are only designed to traverse the latent space, there is no guarantee that their behavior should match our desire. Therefore, our approach should be only treated \textbf{as a proxy} to evaluate the performance of interpolation algorithms. Still, we hope our work can motivate works focusing more on the 
application aspect of interpolation, in addition to distribution matching (also discussed in the next section). 

Next, our evaluation framework requires an inference model $q_\phi(z|x)$ to be available. However, except for some variants such as BiGAN \citep{jeff16bigan}, most GAN models \citep{Goodfellow14gan} do not have an inference network. Thus, our methodology can not be directly applied to GANs.

\vspace{-0.1cm}
\section{Related Works}
\label{sec:related}
\vspace{-0.1cm}
What is the best way to traverse the latent space is an ongoing research topic. \citet{white16spherical} proposes spherical linear interpolation, which treats the interpolation as a great circle path on an n-dimensional hypersphere. Based on optimal transport maps, normalised interpolation \citep{agustsson2018optimal} adapts the latent space operations, so that they fully match the prior distribution, while minimally modifying the original operation. 

With similar motivation, \citet{lesniak2018distributioninterpolation} defines the distribution matching  property (DMP), as a potential guideline for interpolation algorithm design. For standard Gaussian prior, they point out that linear and spherical linear interpolation does not satisfy DMP, while normalised interpolation does. Further, they propose \textit{Cauchy-linear} interpolation, which satisfies DMP for a wide range of prior choices. Finally, \citet{lukasz19realismindex} propose a numerically efficient algorithm that maximizes the \textit{realism index} of an interpolation.

As discussed in the introduction section, most evaluations in the literature are qualitative. \citet{agustsson2018optimal} did a quantitative evaluation, where \textit{inception score} \citep{salimans16ganinception} is used to measure the quality or diversity of decoded samples from the interpolated encodings in an unsupervised manner. Therefore their evaluation focuses more on how well the distribution of the decoded samples matches the data distribution, instead of how ``accurate'' the interpolation algorithm is. In this work since we utilize datasets with attributes, we are able to provide reference $x$ or $z$ for the interpolated points. 

Similar to interpolation, \textit{analogy} \citep{mikolov-etal-2013-linguistic}, usually in the form of 4-tuple $A:B::C:D$ (often spoken as ``A is to B as C is to D''), has been used to demonstrate the structure of the latent space. As pointed out by \citet{agustsson2018optimal}, the most commonly used analogy function $\hat{z}_d:=z_c+(z_b-z_a)$ would also introduce distribution mismatch. With appropriate datasets \citep{reed15analogy}, our evaluation framework can be generalized to compare different latent-space analogy algorithms, and we leave that as future work.

In the disentanglement learning literature \citep{Higgins2017betaVAELB}, several quantitative metrics have been proposed to measure the disentanglement of the latent variable, including z-diff score \citep{Higgins2017betaVAELB}, SAP score \citep{kumar2018variational}, and factor score \citep{pmlr-v80-kim18b}. These metrics are different from our evaluation in that they do not involve traversing the latent space. Also, note that they require datasets with factor annotations (similar to the semantically continuous label used in our work). 

\vspace{-0.1cm}
\paragraph{Related Applications} On the computer vision application side, optimization-based methods~\citep{solomon:hal-01188953} and learning-based methods~\citep{mildenhall2020nerf, Riegler2020FVS, chen2019dibrender, DBLP:journals/corr/abs-1906-07316} have been proposed to synthesize (interpolate) new images and / or 3D shapes from existing observations. Of note, Nerf~\citep{mildenhall2020nerf} enables photorealistic image generation by incorporating neural networks into volumetric rendering. These applications are similar to the image interpolation task considered in this work. However, most of these works did not adopt generative latent variable models.

Related to our application on graph interpolation is the body of works on learning and modeling dynamic graphs, where the goal is mostly to accurately predict future graphs. These works treat a dynamic graph as a sequence of graph snapshots and rely on the combination of recurrent neural networks (RNN) and GNN to perform discrete-time dynamic graph forecasting~\citep{sankar2020dysat,hajiramezanali2019variational,manessi2020dynamic}.
There are RNN-based methods that extracts features from each snapshot using a graph model and then feed them into a Recurrent Neural Network (RNN)~\citep{seo2018structured,taheri2019learning,lei2019gcn,manessi2020dynamic,seo2018structured,chen2018gc,li2019predicting,hajiramezanali2019variational}.
There are also approaches imposing temporal constraints on top of RNN. These constraints include spatial / temporal attention mechanism \citep{sankar2020dysat}, architectural constraints on the static graph model for each snapshot \citep{goyal2018dyngem}, and dynamic parameter constraints (i.e., using another RNN to update parameters) on the static graph model \citep{pareja2020evolvegcn}.

\vspace{-0.1cm}
\section{Conclusion}
\vspace{-0.1cm}
In this work, we show how data labeled with semantically continuous attributes can be utilized to conduct a quantitative evaluation of latent-space interpolation algorithms, for variational autoencoders. Our evaluation can be used to complement the standard qualitative comparison, and also enables evaluation for domains (such as graphs) in which visualization is difficult. 

Interestingly, our experiments reveal that the superiority of interpolation algorithms could be domain-dependent. While normalised interpolation works best in the image domain, spherical linear interpolation achieves the best performance in the graph domain. Next, we test a simple and effective method to restrict the latent space via a bottleneck structure in the encoder. And we report that the interpolation algorithms considered in this work can all benefit from this restriction. Finally, we conduct interpolation-aware training with the labeled attributes, and show that this explicit supervision can greatly boost the interpolation performance.

We also hope this work could motivate future works focusing more on the application potential of latent-space interpolation for deep generative models, in addition to its current role as a demonstration tool of generalization ability.

\bibliography{icml2020}

\newpage
\appendix
\section*{Supplemental Materials}

\section{A: Auxiliary Formulations}
\label{appsec:formulation}

\textbf{Definition of $D_\text{KL}$} For distributions $P$ and $Q$ for continuous random variables, the Kullback–Leibler divergence is defined as:
\begin{equation}
    D_\text{KL}(P||Q) = \int^{+\infty}_{-\infty} p(x) \log \frac{p(x)}{q(x)}.
\end{equation}

\textbf{Normalised spherical interpolation} In section 
Evaluation of Interpolation Algorithms, we attempt to combine $f^\text{slerp}$ and $f^\text{norm}$ as \textit{normalised spherical interpolation}:
\begin{equation}
    f^\text{SN}(z_1, z_2, \lambda) = \frac{\text{sin}[(1-\lambda)\Omega] z_1 + \text{sin}[\lambda \Omega] z_2}{\sqrt{\text{sin}^2[(1-\lambda)\Omega] + \text{sin}^2[\lambda \Omega]}}.
\end{equation}
However, this combination does not bring performance gain.

\section{B: Implementation Details}
\label{appsec:implement_detail}
We report more details of the hyperparameters setting for our evaluations in the following,  

\textbf{ShapeNet} In the unsupervised VAE training, we use a learning rate of $0.0005$, and a minibatch size of $40$. The VAE model is trained for $40000$ iterations. In the interpolation-aware training, we use a learning rate of $0.0005$. For each training iteration, we uniformly sample a new minibatch of $20$ triplets $\{x_{t1},x_{t2},x_{t3}\}$ to optimize $ \mathcal{L}_\text{IAT}$. We use a $\lambda_\text{IAT}$ of $1$ during training. The model is trained for $80000$ iterations.

\textbf{CO3D} In the unsupervised VAE training, we use a learning rate of $0.0001$, and a minibatch size of $100$. The VAE model is trained for $30000$ iterations. In the interpolation-aware training, we use a learning rate of $0.0001$. For each training iteration, we uniformly sample a new minibatch of $40$ triplets $\{x_{t1},x_{t2},x_{t3}\}$ to optimize $ \mathcal{L}_\text{IAT}$. We use a $\lambda_\text{IAT}$ of $1$ during training. The model is trained for $30000$ iterations.

\textbf{Citation Network}
In the unsupervised VAE training, we use a learning rate of $0.0005$, and a minibatch size of $10$. The VAE model is trained for $30000$ iterations. In the interpolation-aware training, we use a learning rate of $0.0005$. For each training iteration, we uniformly sample a new minibatch of $20$ triplets $\{A_{t1},A_{t2},A_{t3}\}$ to optimize $ \mathcal{L}_\text{IAT}$. We use the $\lambda_\text{IAT}$ of $5$ during training. And the model is trained for $50000$ iterations.

\section{C: Auxiliary Experiments}
\label{appsec:auxexp}

Table \ref{tab:co3d_rankdim_metric} shows the results with restricted latent space for CO3D; the results are consistent with ShapeNet and Citation Network: the low-rank encoder benefits latent-space interpolation. Table \ref{tab:co3d_iat} shows the IAT training results for CO3D.

\begin{figure}[h]
  \centering
  \subfloat[ShapeNet]{\includegraphics[width=0.7\linewidth]{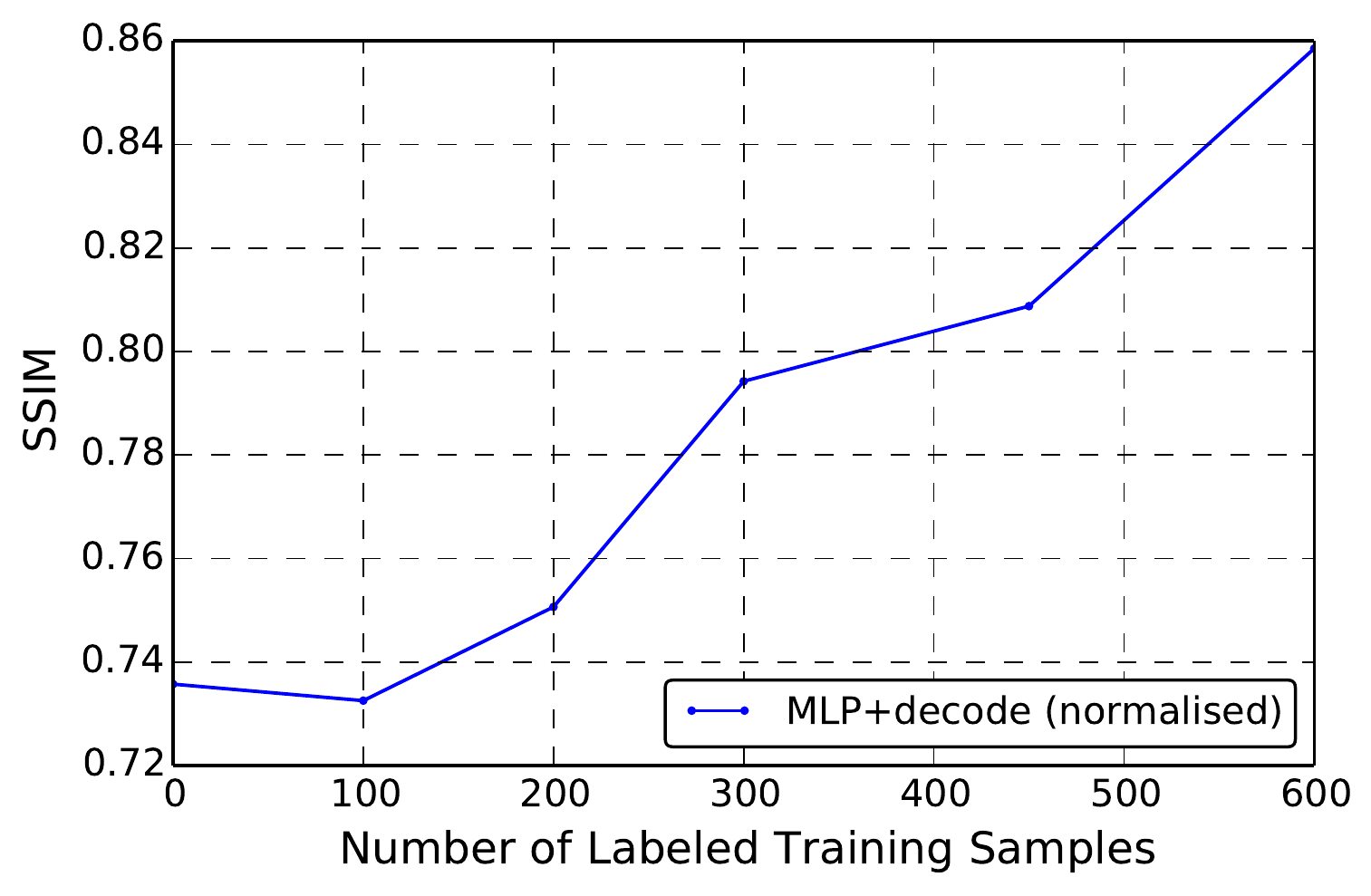}} \\
  \subfloat[Citation Network]{\includegraphics[width=0.7\linewidth]{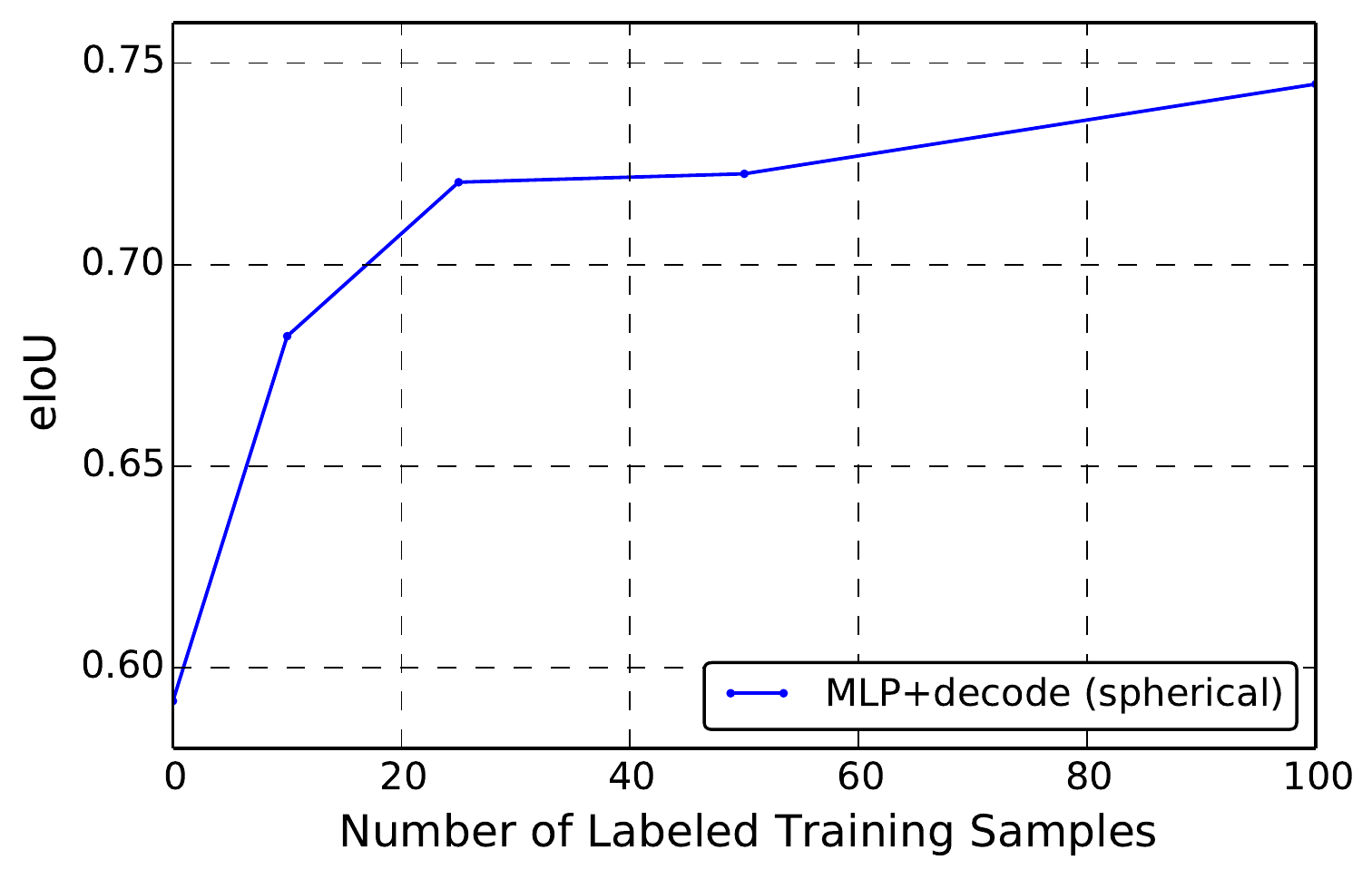}}
  \caption{Performance of IAT with different number of labeled training samples.}
  \vspace{-0.2cm}
  \label{fig:app_difftrainnum}
\end{figure}

\begin{table}[h]
\addtolength{\tabcolsep}{-3.5pt}
\centering
\small
\begin{tabular}{c|ccccc}
\toprule

\multicolumn{6}{c}{\textbf{CO3D ($f^\text{norm}$)}} \\ \midrule
\textbf{IAT} & \multicolumn{1}{c}{\textbf{N/A}} & \multicolumn{1}{c}{\textbf{lat.}} & \multicolumn{1}{c}{\textbf{dec.}} & \multicolumn{1}{c}{\textbf{\begin{tabular}[c]{@{}c@{}}mlp \\ + lat.\end{tabular}}} & \multicolumn{1}{c}{\textbf{\begin{tabular}[c]{@{}c@{}}mlp \\ + dec.\end{tabular}}} \\ \midrule
$\text{MSE}({x}^\text{inter}_{t_2},x_{t_2})\downarrow$  &0.175 &0.174 &0.151 &\textbf{0.131} &0.134\\
$\text{PSNR}({x}^\text{inter}_{t_2},x_{t_2})\uparrow$ &14.33 &14.33 &14.76 &\textbf{15.32} &15.23 \\
\bottomrule
\end{tabular}
\caption{Results of interpolation-aware training (IAT) on CO3D dataset. ``lat.'' refers to $\mathcal{L}^\text{latent}_\text{IAT}$, and ``dec.'' refers to $\mathcal{L}^\text{decode}_\text{IAT}$. The supervision provided by the labeled attributes greatly improve the interpolation performance, comparing to the unsupervised baseline (marked by ``N/A'').}
\vspace{-0.2cm}
\label{tab:co3d_iat}
\end{table}

In Figure \ref{fig:app_difftrainnum}, we provide a study of how the performance improves as the number of labeled data grows. The objective variant is $\mathcal{L}_\text{IAT}^\text{MLP+decode}$. Interestingly, we observe that for the Citation Network dataset, the improvement saturates at around 25 labeled samples, which is around 25\% of the whole training set. 

\begin{table*}
\small
\centering
\footnotesize
\addtolength{\tabcolsep}{-2.0pt}

\begin{tabular}{r|cccccc|cccccc}
\toprule
\textbf{Dataset}                                       & \multicolumn{6}{c|}{\textbf{Latent Dimension}}                               & \multicolumn{6}{c}{\textbf{Encoder Rank}}                                        \\ \midrule
\textbf{CO3D ($f^\text{norm}$)}                    & \textbf{D4019} & \textbf{D1024} & \textbf{D256} & \textbf{D64}     & \textbf{D16} & \textbf{D4} & \textbf{R4096} & \textbf{R1024}  & \textbf{R256}   & \textbf{R64}    & \textbf{R16}  & \textbf{R4}\\ \midrule
$\text{MSE}({x}^\text{inter}_{t_2},x_{t_2})\downarrow$           & 0.174  & 0.176   & 0.178          & 0.181          & 0.187  &0.191  & 0.174  & 0.178         & 0.177        & 0.175           & \textbf{0.170}  &0.178       \\$\text{PSNR}({x}^\text{inter}_{t_2},x_{t_2})\uparrow$ &14.33 &14.32	&14.34	&14.22	&14.18	&13.86 	&14.33 &14.34	&14.32	&14.34	&\textbf{14.37}	&14.02 \\ \midrule
$\text{MSE}({z}^\text{inter}_{t_2},\hat{z}_{t_2})\downarrow$ &0.004 &0.017	&0.071	&0.297	&0.963	&0.803 &0.004 &0.005	&0.005	&0.004	&0.006	&\textbf{0.003}  \\
$\text{cos-sim}({z}^\text{inter}_{t_2},\hat{z}_{t_2})\downarrow$  &0.436 &0.423 &\textbf{0.410}	&0.411	&0.420	&0.527 &0.436 &0.423	&0.425	&0.436	&0.480	&0.555 \\
 \bottomrule
\end{tabular}
\vspace{-0.2cm}
\caption{Interpolation performance with various metrics for low-dimensional ($D$) latent space, or with low-rank ($R$) encoder. The results are from normalized interpolation. For CO3D, we observe a steady PSNR improvement from R4096 to R16.}
\label{tab:co3d_rankdim_metric}
\end{table*}

\end{document}